\documentclass[10pt,twocolumn,letterpaper]{article}

\usepackage{cvpr}
\usepackage{times}
\usepackage{epsfig}
\usepackage{graphicx}
\usepackage{amsmath}
\usepackage{amssymb}

\usepackage{bm}
\usepackage{xcolor}
\usepackage{multirow}
\usepackage{makecell}
\usepackage{textcomp}

\usepackage[pagebackref=true,breaklinks=true,letterpaper=true,colorlinks,bookmarks=false]{hyperref}

\cvprfinalcopy 

\def\cvprPaperID{428} 

\ifcvprfinal\fi

\begin{document}

\title{Modality Shifting Attention Network for Multi-modal Video Question Answering}

\author{Junyeong Kim$^1$\thanks{This work was partly supported by Institute for Information \& communications Technology Planning \& Evaluation(IITP) grant funded by the Korea government(MSIT) (2017-0-01780, The technology development for event recognition/relational reasoning and learning knowledge based system for video understanding)
and partly supported by Institute for Information \& communications Technology Planning \& Evaluation(IITP) grant funded by the Korea government(MSIT) (No. 2019-0-01396, Development of framework for analyzing, detecting, mitigating of bias in AI model and training data)}  \ \ \ \ \ \ Minuk Ma$^1$\ \ \ \ \ \ Trung Pham$^1$\ \ \ \ \ \ Kyungsu Kim$^2$\ \ \ \ \ \ Chang D. Yoo$^1$ \\$^1$ Korea Advanced Institute of Science and Technology (KAIST) \\ $^2$ Samsung Research \\{\tt\small $^1$\{junyeong.kim, akalsdnr, trungpx, cd\_yoo\}@kaist.ac.kr}\ \ \ {\tt\small $^2$ks0326.kim@samsung.com}
}

\maketitle

\begin{abstract}
This paper considers a network referred to as Modality Shifting Attention Network (MSAN) for Multimodal Video Question Answering (MVQA) task. MSAN decomposes the task into two sub-tasks: (1) localization of temporal moment relevant to the question, and (2) accurate prediction of the answer based on the localized moment. The modality required for temporal localization may be different from that for answer prediction, and this ability to shift modality is essential for performing the task. 
To this end, MSAN is based on (1) the moment proposal network (MPN) that attempts to locate the most appropriate temporal moment from each of the modalities, and also on (2) the heterogeneous reasoning network (HRN) that predicts the answer using an attention mechanism on both modalities. MSAN is able to place importance weight on the two modalities for each sub-task using a component referred to as Modality Importance Modulation (MIM). 
Experimental results show that MSAN outperforms previous state-of-the-art by achieving 71.13\% test accuracy on TVQA benchmark dataset. Extensive ablation studies and qualitative analysis are conducted to validate various components of the network.
\end{abstract}

\section{Introduction}
\label{sec:intro}

Bridging the field of computer vision and that of natural language processing appears to be a desiderata of current vision-language tasks.
Sundry efforts that have made progress towards binding the two fields include \cite{fukui-etal-2016-multimodal,videoground1,Liu_2019_ICCV,videoground2} in visual grounding, \cite{Xu_2015_ICML,Venugopalan_2015_ICCV,videocap1,videocap2} in image/video captioning, \cite{Gao_2017_ICCV,Hendricks_2017_ICCV,Yuan_2019_AAAI,2DTAN_2020_AAAI} in video moment retrieval, and \cite{Antol_2015_ICCV,Yang_2016_CVPR,Anderson_2018_CVPR,Gao_2019_CVPR} in visual question answering.
Among the many tasks, VQA is especially challenging as it requires the capability to perform fine-grained reasoning using both image and text. 
This task that requires reasoning has been extended to video question answering (VideoQA) and multi-modal video question answering (MVQA).

\begin{figure}[t]
	\centering
	\includegraphics[width=\columnwidth]{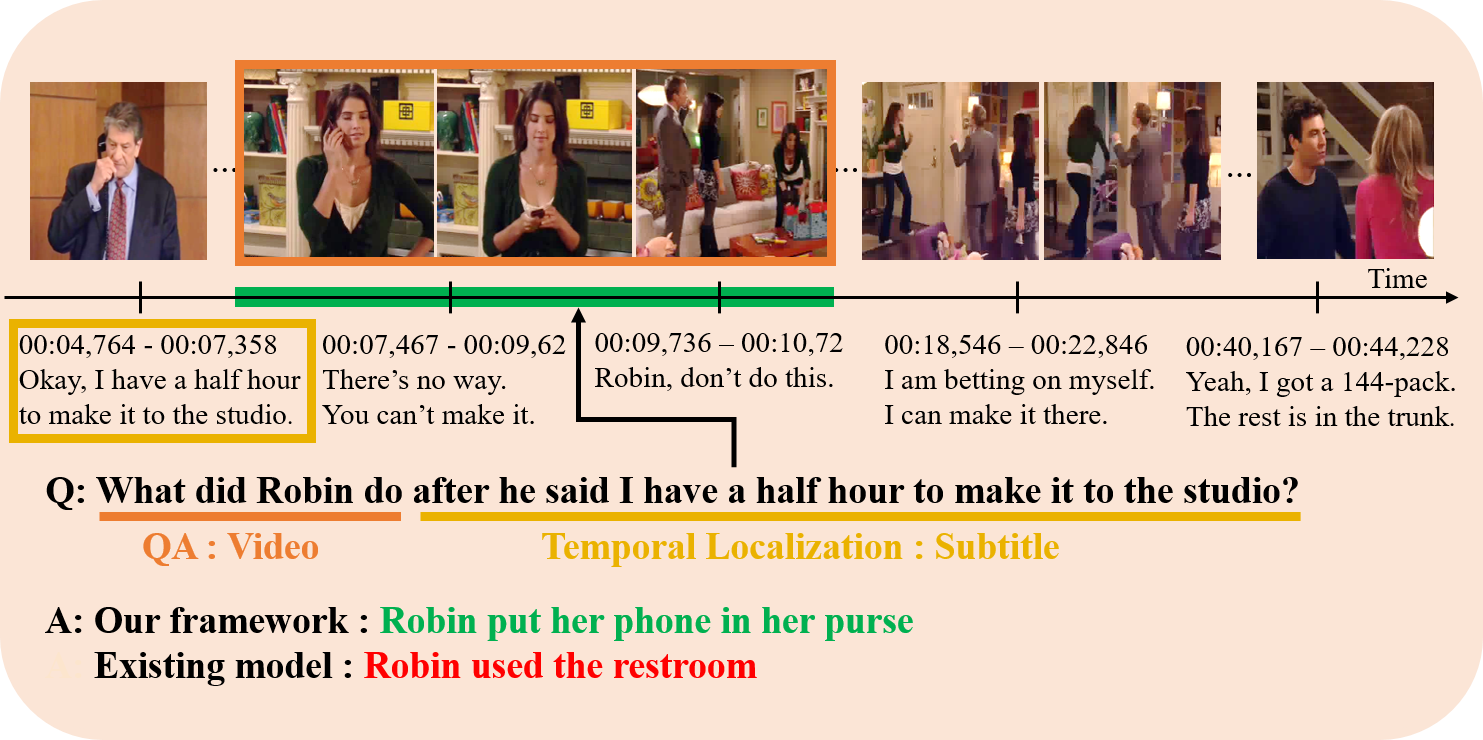}
	\caption{Multimodal Video QA is a challenging task as it requires retrieving the queried information which is interspersed in multiple modalities. For complex question such as \textquotedblleft What did Robin do after he said I have a half hour to make it to the studio?\textquotedblright, we first need to localize the moment by observing subtitle and then infer the answer by looking into video.}
	\label{fig:intro}
\end{figure}

This paper focuses on the task of answering multiple-choice questions regarding a scene in a long untrimmed video based on both video clip and its subtitle. 
This task is referred to as MVQA. 
In comparison to VQA or VideoQA, MVQA is a more challenging task as it (1) requires to locate the temporal moment relevant to the QA, and (2) also requires to perform reasoning on video and subtitle modalities.
To illustrate, consider the question in Fig. \ref{fig:intro} \textit{``What did Robin do after he said I have a half hour to make it to the studio?''}.
To accurately answer the question, the QA system would require the video modality to decipher Robin's action to answer \textit{``What did Robin do''}, and the subtitle modality to localize time-index corresponding to \textit{``after he said ...''}.
%

The first challenge of MVQA is to locate vital moments in all heterogeneous modalities conducive to answering the question.
As \cite{Kim_2019_CVPR} pointed out, the information in the video required to answer the question is not distributed uniformly across the temporal axis.
The temporal attention mechanism has been widely adopted \cite{Tapaswi_2016_CVPR,Na_2017_ICCV,Liang_2018_CVPR,Kim_2018_ECCV,Kim_2019_CVPR} to retrieve relevant information to the question. 
However, it is observed that previous temporal attention are often too blurry or inaccurate in attending important regions of the video and subtitle, and as a result, may introduce itself as noise during the inference.
Aside from qualitatively assessing the predicted attention, until now, no quantitative metric to measure its accuracy was available, which made it difficult to validate the ability to retrieve appropriate information for answering the question.
%

The second challenge of MVQA is to be capable of reasoning on heterogeneous modalities for answering the question.
Early studies on MVQA adopted an early-fusion framework \cite{Kim_2017_IJCAI,Na_2017_ICCV} that fuse video and subtitle into a joint embedding space at the early stage of the prediction pipeline which thereafter is the basis for subsequent reasoning followed by final prediction.
Recent methods are based on the late-fusion framework \cite{Kim_2018_ECCV,Lei_2018_EMNLP,Kim_2019_CVPR}, which independently process video and subtitle and then combine the two processed outputs for final prediction.
The two extreme frameworks have their upsides as well as downsides.
The early-fusion framework can be extremely useful for moment location as well as for performing reasoning for answer prediction only when the sample space is well populated such that joint embedding space is well defined; otherwise, extreme overfitting can occur, and one modality will act as noise on the other modality.
The late-fusion framework is often inadequate for answering questions that require one modality for temporal localization and another for answer prediction as shown in the example Fig. \ref{fig:intro}.
We think such \textit{Modality Shifting Ability} is an essential component of MVQA, which existing methods are incapable of.

To resolve the aforementioned challenges, we first propose to decompose the problem of MVQA into two sub-tasks: temporal moment localization and answer prediction.
The key motivation of this paper comes from the fact that the modality required for temporal moment localization may be different from that required for answer prediction.
To this end, Modality Shifting Attention Network (MSAN) is proposed with the following two components: (1) moment proposal network (MPN) and (2) heterogeneous reasoning network (HRN).
MPN localizes the temporal moment of interest (MoI) that is required for answering the question.
Here, the MoI candidates are defined over both video and subtitle, and MPN learns the moment scores for each MoI candidate.
Based on the localized MoI, HRN infers the correct answer through a multi-modal attention mechanism called Heterogeneous Attention Mechanism (HAM). 
HAM is composed of three attention units: self-attention (SA) that models the intra-modality interactions (i.e., word-to-word, object-to-object relationships), context-to-query (C2Q) attention that models the inter-modality interactions between question and context (i.e., video and subtitle), and context-to-context (C2C) attention to model the inter-modality interactions between video and subtitle.
The results of MPN and HRN are further adjusted by Modality Importance Modulation (MIM) which is an additional attention mechanism over modalities.

%
%
%

\section{Related Works}

\subsection{Visual Question Answering}
Visual Question Answering (VQA) \cite{Antol_2015_ICCV} aims at inferring the correct answer of a given question regarding the visual contents in an image.
Yang \textit{et al.} \cite{Yang_2016_CVPR} proposed stacked attention mechanism which performs multi-step reasoning by repeatedly attending relevant image regions, and refines the query after each reasoning step.
Anderson \textit{et al.} \cite{Anderson_2018_CVPR} introduced extracting object proposals in the image using Faster R-CNN \cite{Ren_2016_PAMI} and the question is used to attend to the proposals.
DFAF \cite{Gao_2019_CVPR} utilizes both self- and co-attention mechanism to dynamically fuse multi-modal representations with intra- and inter-modality information flows.
Video Question Answering (VideoQA) \cite{Zhou_2017_IJCV,Jang_2017_CVPR} is a natural extension of VQA into the video domain.
Jang \textit{et al.} \cite{Jang_2017_CVPR} extracted both the appearance feature and motion features as visual representation, and used spatial and temporal attention mechanism to attend to the moments in video and the regions in frames.
Co-memory attention \cite{Gao_2018_CVPR} contains two separate memory modules each for appearance and motion cues, and each memory guides the other memory while generating the attention.
Fan \textit{et al.} \cite{Fan_2019_CVPR} proposed heterogeneous video memory to capture global context from both appearance and motion features, and question memory to understand high-level semantics in question.

\begin{figure*}[t]
	\centering
	\includegraphics[width=\textwidth]{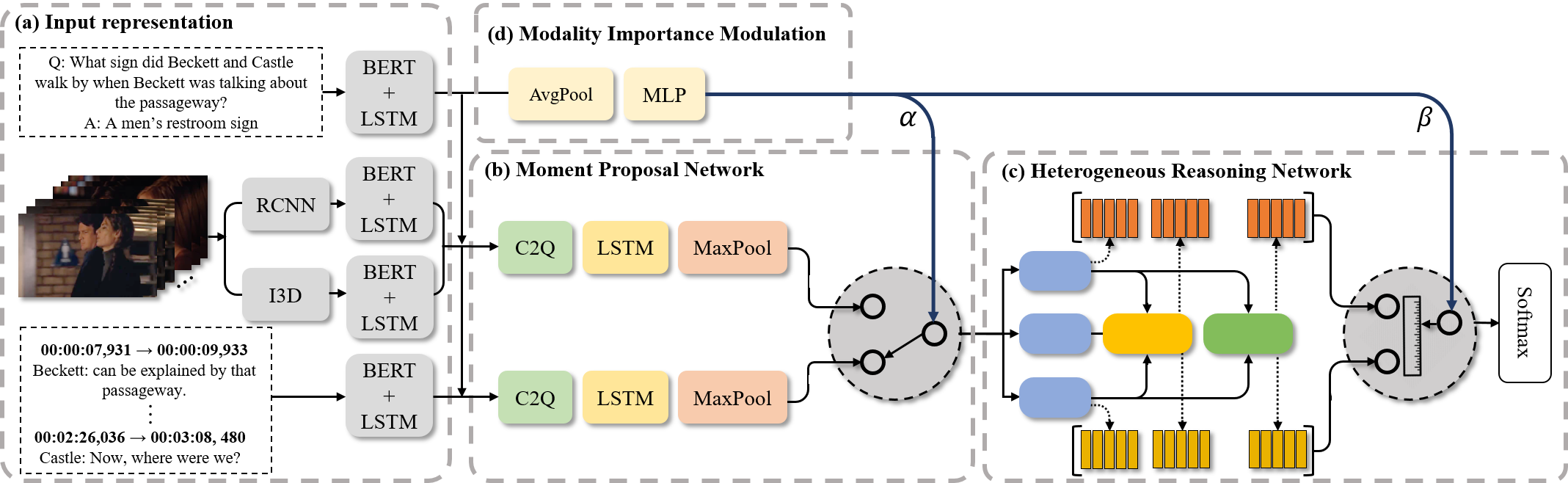}
	\caption{Illustration of modality shifting attention network (MSAN) which is composed of the following components: (a) Video and text representation utilizing BERT for embedding, (b) Moment proposal network to localize the required temporal moment of interest for answering the question, (c) Heterogeneous reasoning network to infer the correct answer based on the localized moment, and (d) Modality importance modulation to weight the output of (b) and of (c) differently according to their importance.}
	\label{fig:overall}
\end{figure*}

\subsection{Multi-modal Video Question Answering}
Multi-modal Video Question Answering (MVQA) further extends VideoQA to leverage text modality, such as a subtitle, in addition to video modality.
The inclusion of text modality makes the reasoning more challenging as the vital information required to answer the question is interspersed in both video and text modality. 
In the early stage of MVQA research, early-fusion was commonly used to fuse multiple modalities. 
Na \textit{et al.} \cite{Na_2017_ICCV} proposed a read-write memory network (RWMN) which utilizes a CNN-based memory network to write and read the information to and from memory.
As video conveys a fairly different context compared to the subtitle, early-fusion may produce noise at feature-level and interfere with retrieving semantic context.
To this end, recent methods \cite{Kim_2018_ECCV,Kim_2019_CVPR,Lei_2018_EMNLP,Kim_2019_IJCNN} took late-fusion approaches to merge multiple modalities.
The two-stream network \cite{Lei_2018_EMNLP} provides a simple late-fusion method with a bi-LSTM context encoder followed by context-to-query attention mechanism.
Multi-task Learning (MTL) \cite{Kim_2019_IJCNN} further extends the two-stream network by leveraging modality alignment and temporal localization as additional tasks. 
Progressive Attention Memory Network (PAMN) \cite{Kim_2019_CVPR} utilizes QA pairs to temporally attend video and subtitle memories, and merge using a soft attention mechanism.

\section{Modality Shifting Attention Network}
\label{sec:msan}
Figure \ref{fig:overall} shows the overall pipeline of modality shifting attention network (MSAN) with two sub-networks: Moment Proposal Network (MPN) and Heterogeneous Reasoning Network (HRN).
The main focus of MSAN comes from the observation that the reasoning in MVQA can be accomplished by two consecutive sub-tasks: (1) temporal moment localization, and (2) answer prediction and that each sub-task may require different modality more than the other.

\subsection{Input Representation}
\label{ssec:feature}

\textbf{Video Representation.} 
The input video is represented as a set of detected object labels (i.e. visual concepts) as in other recent methods on MVQA \cite{Lei_2018_EMNLP,Kim_2019_IJCNN,Kim_2019_CVPR}.
Specifically, the video was sampled at 3 FPS to form set of frames $\{v_t\}_{t=1}^{F}$ where $F$ is the number of frames. 
Then Faster R-CNN \cite{Ren_2016_PAMI} pre-trained on Visual Genome benchmark \cite{Krishna_2017_IJCV} is used to detect visual concepts composed of object label and its attribute (e.g. \textit{gray pants}, \textit{blue sweater}, \textit{brown hair}, etc).
We divide the input video into a set of video \textit{shots} to remove redundancy.
When a scene is not changing fast, the visual concepts in nearby frames may be redundant. 
We define video \textit{shot} as the set of successive frames whose intersection over union (IoU) of visual concepts is more than 0.3. 
The input video is divided into video \textit{shots} in chronological order for removing duplicate concepts. 
In contrast to video, we do not define \textit{shots} for the subtitle as it is assumed there is little redundancy in the conversation.
Inspired by VideoQA \cite{Jang_2017_CVPR,Fan_2019_CVPR}, we also incorporate motion cues in our framework.
To our knowledge, while none of the existing methods on MVQA utilize motion cues, we observed that motion cues might help understanding video clip to answer the question. 
For each video shot generated above, I3D \cite{Carreira_2017_CVPR} pre-trained on Kinetics benchmark \cite{Carreira_2017_CVPR} is used to produce top-5 action labels which we refer to as \textit{action concept}.
Visual and action concepts are concatenated to represent the corresponding video shot.
As visual and action concepts are in the text domain, they are embedded in the manner as the subtitle. 

\textbf{Text Representation.}
%
%
We extracted 768-dimensional word-level text representations for shots in a video, sentences in the subtitle, and QA pairs from the second-to-last layer of BERT-Base model \cite{Devlin_2018_ArXiv}. 
The extracted representations were fixed during training.
The question and each of answer candidates were concatenated to form five hypothesis $\{h_k\}_{k=1}^{5}$ where $h_k \in \mathbb{R}^{n_{h_k} \times 768}$ and $n_{h_k}$ represents the number of words in $k^{th}$ hypothesis.
For each hypothesis, MSAN learns to predict its correctness score and to maximize the score of the correct answer.
For simplicity, we drop subscript $k$ for the hypothesis in the following sections.

\subsection{Moment Proposal Network}
\label{ssec:mpn}
Moment Proposal Network (MPN) localizes the required temporal moment of interest (MoI) for answering the question.
The MoI candidates are generated for temporally-aligned video and subtitle.
For each MoI candidate, MPN produces two moment scores, one for each modality. 
The Modality Importance Modulation (MIM) adjusts the moment score of each modality to weight on the important modality for temporal moment localization.
MPN is trained to maximize the scores of the positive MoIs using ranking loss. 

\subsubsection{Moment of Interest Candidate Generation}
\label{ssec:MoI}

We generate $N$ moments of interest (MoI) candidates for temporally-aligned video and subtitle using pre-defined sliding windows.
Each MoI candidate consists of a set of video shots and subtitle sentences which is flattened and represented as $v \in \mathbb{R}^{n_v \times 768}, s \in \mathbb{R}^{n_s \times 768}$, respectively. 
Here, $n_v$ is the number of visual objects in video, and $n_s$ is the number of words in the subtitle.
We defined various lengths of sliding windows for each modality so that the MoI candidates are distributed evenly along the temporal axis and cover the entire video. 
We label the MoI candidate as positive if it has $IoU \geq 0.5$ with the provided GT moment, and the other MoI candidates are labeled as negatives.  
We obtain the final features $V, S, H$ by passing the BERT embeddings $v, s, h$ through one-layer bi-directional LSTM network.

\subsubsection{MoI Candidate Moment Score}
Among $N$ MoI candidates, MPN localizes the relevant MoI for answering the question.
MPN first produces video/subtitle moment scores for each MoI candidate.
We first utilize context-to-query (C2Q) attention to jointly model each context (i.e., video, subtitle) and the hypothesis and obtain $V^H$ and $S^H$.
Details of C2Q attention can be found in following Sec. \ref{sssec:ham}.
Then, we feed the concatenated features $[V;V^H]$ and $[S;S^H]$ into one-layer bi-directional LSTM followed by max-pooling along the temporal axis.
The final video and subtitle features $f^v, f^s \in \mathbb{R}^{d}$ are passed through shared score regressor (FC($d$)-ReLU-FC(1)-$\sigma$) that outputs the video/subtitle moment scores $m^v, m^s$ for video and subtitle, respectively.

\subsubsection{Modality Importance Modulation}
\label{sssec:mim1}
To place more weight on important modality for temporal moment localization, the moment scores are adjusted by Modality Importance Modulation (MIM). 
The moment scores of the important modality are boosted while those of the counterpart are suppressed. 
The coefficient $\alpha$ used for the modulation is obtained by passing average pooled question into an MLP (FC($d$)-ReLU-FC(1)) with sigmoid activation to constrain the range of $\alpha$.
MIM is formulated as:
\begin{eqnarray}
\alpha &=& \sigma(MLP(q)), \\
m^v &\leftarrow& F_M(m^v, \alpha), \\
m^s &\leftarrow& F_M(m^s, 1-\alpha),
\end{eqnarray}
where $F_M$ is the modulation function. We consider three types of modulation functions: additive $F_M^{\mbox{a}}$, multiplicative $F_M^{\mbox{m}}$, and residual $F_M^{\mbox{r}}$:
\begin{eqnarray}
	F_M^{\mbox{a}}(m_i, \alpha) &=& m_i+\alpha, \\
	F_M^{\mbox{m}}(m_i, \alpha) &=& m_i \cdot \alpha, \\
	F_M^{\mbox{r}}(m_i, \alpha) &=& m_i+m_i \cdot \alpha.
\end{eqnarray}
During inference, MPN selects the MoI candidate with the largest moment score for answer prediction.
\textbf{cross-modal ranking loss} is proposed to train MPN, which encourages the moment scores of the positives MoI candidate to be greater than the negatives by a certain margin. 
Rather than applying the ranking loss on each modality, we propose to aggregate the moment scores from both modalities and apply the ranking loss. 
We call this cross-modal ranking loss $\mathcal{L}_{cmr}$ which is represented as follows:
\begin{equation}
\begin{split}
\mathcal{L}_{cmr} &= \sum_{p^+, p^- \in p} \mathcal{L}^R(p^+, p^-), 
\end{split}
\end{equation}
where $p+,p-$ denotes the scores of positive and negative candidate moments respectively, and $\mathcal{L}^R(x,y)=\mbox{max}(0, x-y+b)$ is the ranking loss with margin $b$. During training, we sampled the same number of positives and negatives for stable learning.

\textbf{Relationship between MPN and Other Methods} 
The main philosophy behind MPN is similar to the region proposal network (RPN) \cite{Ren_2016_PAMI}, which is widely used for object detection. 
While RPN defines a set of anchors along the spatial dimension, MPN defines a set of MoI candidates along the temporal dimension. 
In both cases, the end-classifier is trained that takes the detected feature as input and outputs an object class or the index of the correct answer.
However, MPN is a conditional method in that the behavior changes conditioned on the input question. 
As MPN localizes a specific temporal region, it can be seen as a type of hard attention mechanism. 
In contrast to soft temporal attention mechanism, which has been the dominant mechanism in previous works, we believe that MPN is more intuitive, measurable by fair metrics, and less noisy.

\begin{figure}[t]
	\centering
	\includegraphics[width=7cm]{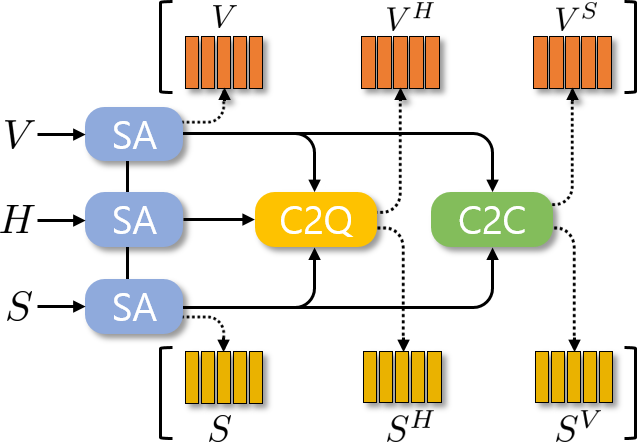}
	\caption{Illustraction of Heterogeneous Attention Mechanism with three attention units; self-attention (SA), context-to-query (C2Q) attention, and context-to-context (C2C) attention. }
	\label{fig:hrn}
\end{figure}

\subsection{Heterogeneous Reasoning Network}
\label{ssec:hrn}
Heterogeneous Reasoning Network (HRN) takes the localized MoI by MPN and learns to infer the correct answer.
HRN involves parameter-efficient heterogeneous attention mechanism (HAM) to consider inter- and intra-modality interactions of heterogeneous modalities.
HAM enables rich feature interactions by transforming the video and subtitle features by representing each element in video or subtitle in all three heterogeneous modality feature spaces.
The Modality Importance Modulation (MIM) again modulates the output of HRN to weight on the important modality for answer prediction.

\subsubsection{Heterogeneous Attention Mechanism}
\label{sssec:ham}
Heterogeneous attention mechanism (HAM) is introduced to consider the inter- and intra- modality interactions by representing a feature in one modality by the linear combination of the features of the other modalities. 
HAM is composed of three basic attentional units: self-attention (SA), context-to-query (C2Q) attention, and context-to-context (C2C) attention, all of which are based on the \textit{dot product attention}. 
%

For two sets of input features $X \in \mathbb{R}^{m \times d}$ and $Y \in \mathbb{R}^{n \times d}$, dot-product attention first evaluates the dot-product of every element of $X$ and $Y$ in obtaining a similarity matrix.
Then softmax function is applied on each row of the similarity matrix in obtaining an attention matrix of size $m \times n$.
The attended feature $X^Y$ is obtained by multiplying attention matrix and $Y$:
\begin{equation}
X^Y = A(X, Y) = \mbox{softmax}(XY)Y.
\end{equation}
We can interpret the dot-product attention as describing each element $x_i$ of $X$ in the feature space of $Y$ by representing $x_i$ with a linear combination of elements in $Y$ with respect to cross-modal similarity.
%

The self-attention (SA) unit is the dot-product attention of feature with itself for defining the intra-modality relations. 
SA unit is represented as $A(X, X)$ where $X$ is an input feature. 
The C2Q and C2C attention units consider the inter-modality relationships and defined as: $A(C,Q)$ and $A(C,C)$, respectively. 
The three attentional units are combined in a modular way in defining the Heterogeneous Attention Mechanism, as illustrated in Figure \ref{fig:hrn}. In HRN, HAM takes the localized video $V$, subtitle $S$, hypothesis $H$ as inputs, and outputs two transformed context features $\tilde{V}, \tilde{S}$. First, each feature is updated by SA units. Then, the context is transformed into the hypothesis space by C2Q unit and the other context space by C2C unit as described mathematically below:
\begin{eqnarray}
&V \leftarrow A(V,V), S \leftarrow A(S,S), H \leftarrow A(H,H),\\
&V^H = A(V,H), S^H = A(S,H), \\
&V^S = A(V,S), S^V = A(S,V).
\end{eqnarray}
Finally, we concatenate the output of three units along feature dimension to construct the rich context descriptor as described below:
\begin{eqnarray}
\tilde{V} &=& [V; V^H; V^S] \in \mathbb{R}^{n_v \times 3d}, \\
\tilde{S} &=& [S; S^H;\ S^V] \in \mathbb{R}^{n_s \times 3d}.
\end{eqnarray}
As a consequence, $\tilde{V}$ is represented as a concatenation of itself in the video feature space, hypothesis feature space, and subtitle feature space while $\tilde{S}$ is the representation of the subtitle as a concatenation of itself in three feature spaces: subtitle, hypothesis, and video.

\textbf{Relationships of HAM to Other Methods}
Recent studies in VQA \cite{Gao_2019_CVPR,Yu_2019_CVPR} have shown that simultaneously learning self-attention and co-attention for visual and textual modalities leads to a more accurate prediction.
Inspired by the previous works on self-attention and co-attention, HAM combines three attentional units to achieve temporal multi-modal reasoning by rich feature interactions between the video, subtitle, and hypothesis.
Also, while previous co-attention \cite{Yu_2019_CVPR} is more about highlighting the important features, the attentional units of HAM perform feature transform from one space to another space. 
While multi-head attention \cite{Vaswani_2017_NIPS} is widely adopted in VQA, the number of parameters is prohibitively large for MVQA, where there are more than a few hundred of objects and words in video and subtitle. 

\subsubsection{Modality Importance Modulation and Answer Reasoning}
With the heterogeneous attention learning, the output video feature $\tilde{V} \in \mathbb{R}^{n_v \times 3d}$ and subtitle feature $\tilde{S} \in \mathbb{R}^{n_s \times 3d}$ contain information rich with regards to various modalities.
The heterogeneous representations of the video $\tilde{V}$ and subtitle $\tilde{S}$ are fed into a one-layer bi-directional LSTM and max-pooling along temporal axis to form final feature vectors.
We utilize two-layer MLP (FC($d$)-ReLU-FC($5$)) to obtain the prediction scores $\ell^v, \ell^s \in \mathbb{R}^5$ for each video and subtitle.
Again, the prediction scores $\ell^v$ and $\ell^s$ are adjusted by Modality Importance Modulation (MIM):
\begin{eqnarray}
\beta &=& \sigma(MLP(q)), \\
\ell &=& \beta \ell^v + (1 - \beta)\ell^s,
\end{eqnarray}
where $\ell$ represents the final prediction score.
We use standard cross-entropy (CE) as the loss function to train 5-way classifier on top of the final prediction score $\ell$.

\section{Experiments}

\subsection{Datasets}
TVQA \cite{Lei_2018_EMNLP} dataset is the largest MVQA benchmark dataset.
TVQA dataset contains human annotated multiple-choice question-answer pairs for short video clips segmented from 6 long running TV shows: \textit{The Big Bang Theory, How I Met Your Mother, Friends, Grey's Anatomy, House, Castle}.
The questions in TVQA are formatted as follows: \textquotedblleft[What/How/Where/Why/...] \underline{\ \ \ \ \ \ \ \ \ \ \ \ } [when/before/after] \underline{\ \ \ \ \ \ \ \ \ \ \ \ }?\textquotedblright.  
The second part of the question localizes the relevant moment in the video clip, and the first part asks question about localized moment.
Each question contains 5 answer candidates and only one of them is correct.
There are total 152.5K QA pairs and 21,793 video clips in TVQA which splits into 122,039 QAs from 17,435 clips for train set, 15,252 QAs from 2.179 clips for validation set and 7,623 QAs from 1,089 clips for test set, respectively.

\subsection{Experimental Details}
%
%

%
The entire framework is implemented with PyTorch \cite{Paszke_2017_NIPSw} framework.
We set the batch size to 16.
Adam optimizer \cite{Kingma_2014_ICLR} is used to optimize the network with the initial learning rate of 0.0003.
All of the experiments were conducted using NVIDIA TITAN Xp (12GB of memory) GPU with CUDA acceleration.
We trained the network up to 10 epochs with early stopping in the case of validation accuracy doesn't increase for 2 epochs.
In all the experiments, recommended train / validation / test split was strictly followed. 

\subsection{Ablation Studies}

\subsubsection{Ablation Study on Moment Proposal Network}
\label{sssec:mpn_result}
This sections describes the quantitative ablation study on Moment Proposal Network (MPN).
Given two temporal moments $(s_1,e_1),(s_2,e_2)$, the Intersection over Union (IoU) is defined by:
\begin{equation}
\mbox{IoU}=1.0*\frac{\mbox{min}(e_1,e_2)-\mbox{max}(s_1,s_2)}{\mbox{max}(e_1,e_2)-\mbox{min}(s_1,s_2)}.
\end{equation}
The gist of MPN is to prune out irrelevant temporal regions. 
Therefore, it is preferable that the localized MoI overlaps with the ground truth.
To reflect such preference, the Coverage metric is proposed which is represented as:
\begin{equation}
\mbox{Cov}=1.0*\frac{\mbox{min}(e_1,e_2)-\mbox{max}(s_1,s_2)}{e_2-s_2}.
\end{equation}
Table \ref{tab:mpn} summarizes the quantitative ablation study on MPN. 
Without Modality Importance Modulation, MPN still can rank the MoI candidates to some extent due to the cross-modal ranking loss.
Three modulation functions enhanced the quality of MPN by \texttildelow 6.0\% of IoU. 
Even the best candidate moment may not perfectly overlap with the ground truth. 
Therefore, we also introduced some safety margin by expanding the temporal boundaries of the predict moment during inference. 
This lowers the IoU, but increases the coverage which helps to include the ground truth moment.  

\begin{table}[t]
	\centering
	\caption{Ablation study on Moment Proposal Network (MPN).}\smallskip
	\begin{tabular}{l||c c}
		\Xhline{3\arrayrulewidth}
		Method                       & IoU  & Cov  \\ \Xhline{3\arrayrulewidth}
		additive w/o MIM             & 0.25 & 0.32 \\ 
		additive                     & 0.29 & 0.52 \\
		multiplicative               & 0.31 & 0.54 \\ 
		residual                     & 0.30 & 0.54 \\ \hline
		ideal                        & 0.76 & 1    \\ \Xhline{3\arrayrulewidth}
	\end{tabular}
	\label{tab:mpn}
\end{table}

\begin{table}[t]
	\centering
	\caption{Ablation study on model variants of MSAN on the validation set of TVQA. The last column shows the performance drop compared to the full model of MSAN.}\smallskip
	\begin{tabular}{l||c c}
		\Xhline{3\arrayrulewidth}
		Methods                    & valid Acc.        & $\Delta$    \\ \Xhline{3\arrayrulewidth}
		MSAN w/o MPN               &  69.89            & -0.9\%      \\
		MSAN w/ GT moment          &  71.62            & +0.83\%     \\ \hline  
		MSAN w/o SA                &  70.21            & -0.58\%     \\
		MSAN w/o C2C               &  70.47            & -0.32\%     \\ \hline
		MSAN w/o MIM on MPN        &  70.56            & -0.23\%     \\
		MSAN w/o MIM on HRN        &  70.35            & -0.44\%     \\ \hline 
		MSAN                       &   70.79           &    0        \\ \Xhline{3\arrayrulewidth}
		
	\end{tabular}
	\label{tab:abl_variant}
\end{table}

\subsubsection{Ablation study on Model Variants}
Table \ref{tab:abl_variant} summarizes the ablation analysis on model variants of MSAN on the validation set of TVQA in order to measure the effectiveness of the proposed key components.
The first block of Table \ref{tab:abl_variant} provides the ablation results of MPN to the overall performance.
Without MPN (i.e. using the full video and subtitle), the accuracy is 69.89\%.
When the ground truth MoI is given, the accuracy is 71.62\%.
With MPN, the overall accuracy is 70.79\% which is 0.90\% higher than the MSAN w/o MPN. 
The second block of Table \ref{tab:abl_variant} provides the ablation results on HRN.
Without SA, there is a 0.58\% of performance drop.
Without C2C attention, there is a 0.32\% of performance drop.
The third block of Table \ref{tab:abl_variant} provides the ablation results on MIM.
Without the MIM on MPN (i.e. the moment score by MPN is not modulated), there is a 0.23\% of performance drop.
Without the MIM of HRN (i.e. the video/subtitle logits from HRN are summed instead of weighting), there is a 0.44\% of performance drop. 
Therefore MIM increases the overall performance. 
MIM also contributes to interpreting the inference of the model by suggesting what modality was more important to retrieve the moment.

\subsubsection{Comparison with the state-of-the-art methods}
Table \ref{tab:acc_tvqa} summarizes the experimental results on TVQA dataset.
We compare with the state-of-the-art methods two-stream \cite{Lei_2018_EMNLP}, PAMN \cite{Kim_2019_CVPR} and MTL \cite{Kim_2019_IJCNN} and performances reported to online evaluation server (i.e. ZGF and STAGE).
The ground-truth answers for TVQA test set are not available and test set evaluation can only be performed through an online evaluation server.
MSAN achieves the test accuracy of $71.13\%$ which outperforms the previously best method by $4.08\%$, establishing the new state-of-the-art.
%

For fair comparison with the previous methods with respect to feature representation, we also provide the results of MSAN using ImageNet feature and GloVe \cite{Pennington_2014_EMNLP} text representation.
The provided results consistently indicate that our MSAN outperforms current state-of-the-art methods by achieving the performance of $68.18\%.$
%
While none of the currnet MVQA methods make use of motion cues, we extracted action concept representation from video clip and provide the results using it.
Compared to MSAN with vcpt (70.92\%), incorporating motion cues provides 0.21\% performance gain.

\begin{table}[t]
	\centering
	\caption{Comparison with the state-of-the-art method on TVQA dataset. \textquotedblleft img\textquotedblright\ is imagenet feature, \textquotedblleft reg\textquotedblright\ is regional feature and \textquotedblleft vcpt\textquotedblright\ is visual concept feature and \textquotedblleft acpt\textquotedblright is action concept feature.}\smallskip
	\begin{tabular}{l||c c c}
		\Xhline{3\arrayrulewidth}
		Methods                                          &     Text Feat.  & Video Feat. & test Acc.       \\ 
		\Xhline{3\arrayrulewidth}
		\multirow{3}{*}{two-stream \cite{Lei_2018_EMNLP}}   & \multirow{3}{*}{GloVe}           & img         & 63.44           \\
		& & reg         & 63.06           \\
		& & vcpt         & 66.46           \\ \hline
		\multirow{2}{*}{PAMN \cite{Kim_2019_CVPR}}        & \multirow{2}{*}{GloVe}     & img         & 64.61           \\
		& & vcpt         & 66.77           \\ \hline
		\multirow{2}{*}{MTL \cite{Kim_2019_IJCNN}}        &  \multirow{2}{*}{GloVe}    & img         & 64.53           \\
		& & vcpt         & 67.05           \\ \hline
		ZGF                                               & - & - & 68.90 \\
		STAGE \cite{STAGE}                                            & BERT & reg & 70.23 \\ \hline
		\multirow{1}{*}{MSAN}                      & GloVe     & vcpt        & 68.18           \\
		\multirow{3}{*}{MSAN}
		&\multirow{3}{*}{BERT} & vcpt        & 70.92           \\
		& & acpt        & 68.57           \\
		& & vcpt+acpt   & \textbf{71.13}  \\
		\Xhline{3\arrayrulewidth}
	\end{tabular}
	\label{tab:acc_tvqa}
\end{table}

\subsection{Qualitative Analysis}
\subsubsection{Performance by question type}
\begin{figure}[h]
	\centering
	\includegraphics[width=\columnwidth]{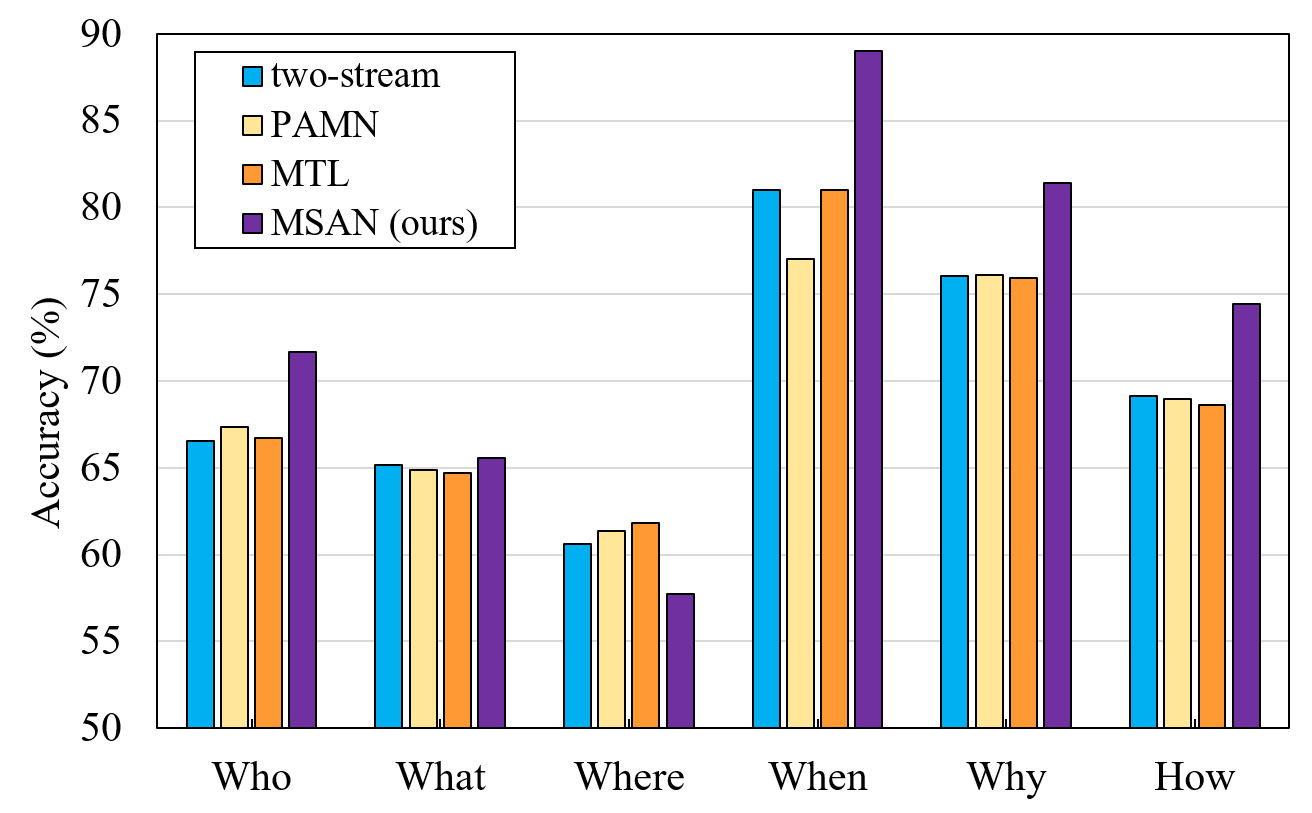}
	\caption{Performance of two-stream, PAMN, MTL, and MSAN by question type on TVQA validation set.}
	\label{fig:qual2}
\end{figure}
We further investigate the performance of MSAN by comparing the accuracy with respect to question type.
Figure \ref{fig:qual2} shows performance comparison by question type on TVQA validation set.
We divided the question types based on 5W1H (i.e. \textit{Who, What, Where, When, Why, How}).
For fair comparison with existing methods, we first tried to reproduce the results on two-stream, PAMN, MTL and obtained the following validation performances; 66.39\%, 66.38\%, 66.22\%, respectively. 
For the majority of question types, MSAN shows significantly better performance than the others.
Especially, MSAN achieves 89\% on \textquotedblleft when\textquotedblright\ question.

\subsubsection{Analysis by question type and required modality}
This sections describes the analysis of MSAN by the question type and the required modality by each question. 
For this, we labeled \texttildelow 5000 samples in the validation set of TVQA according to which modality is required for temporal moment localization and which modality is required for answer prediction.
For example, the label for the question \textit{``What did Phoebe say after the group hug?''} is $(S, V)$ as it asks `say' (i.e. subtitle) and indicates the moment of `group hug' (i.e. video). In this way, there are four types of labels: $(S,S),(S,V),(V,S),(V,V)$. 

\begin{figure}[h]
	\centering
	\includegraphics[width=\columnwidth]{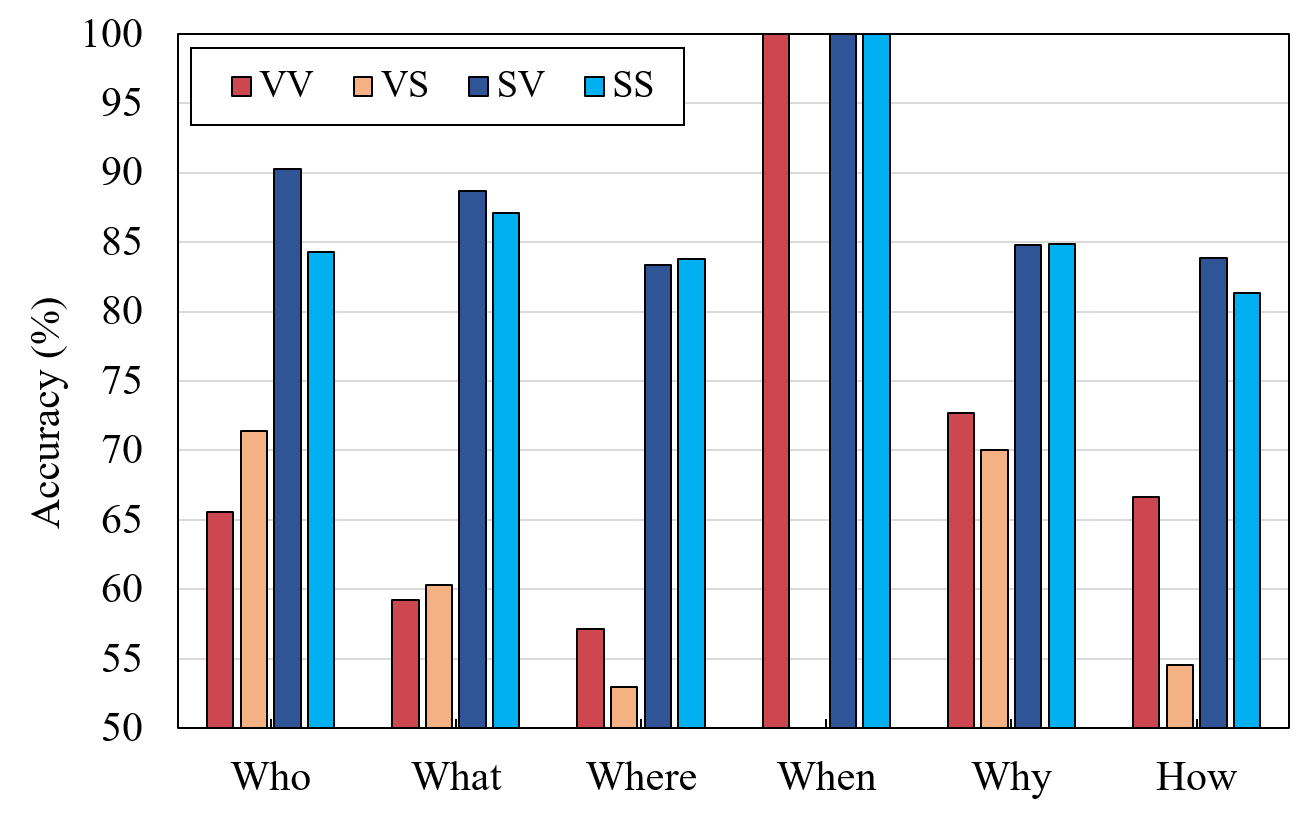}
	\caption{Analysis by question type and required modality of MSAN}
	\label{fig:modality}
\end{figure}

\begin{figure*}[t]
	\centering
	\includegraphics[width=\textwidth]{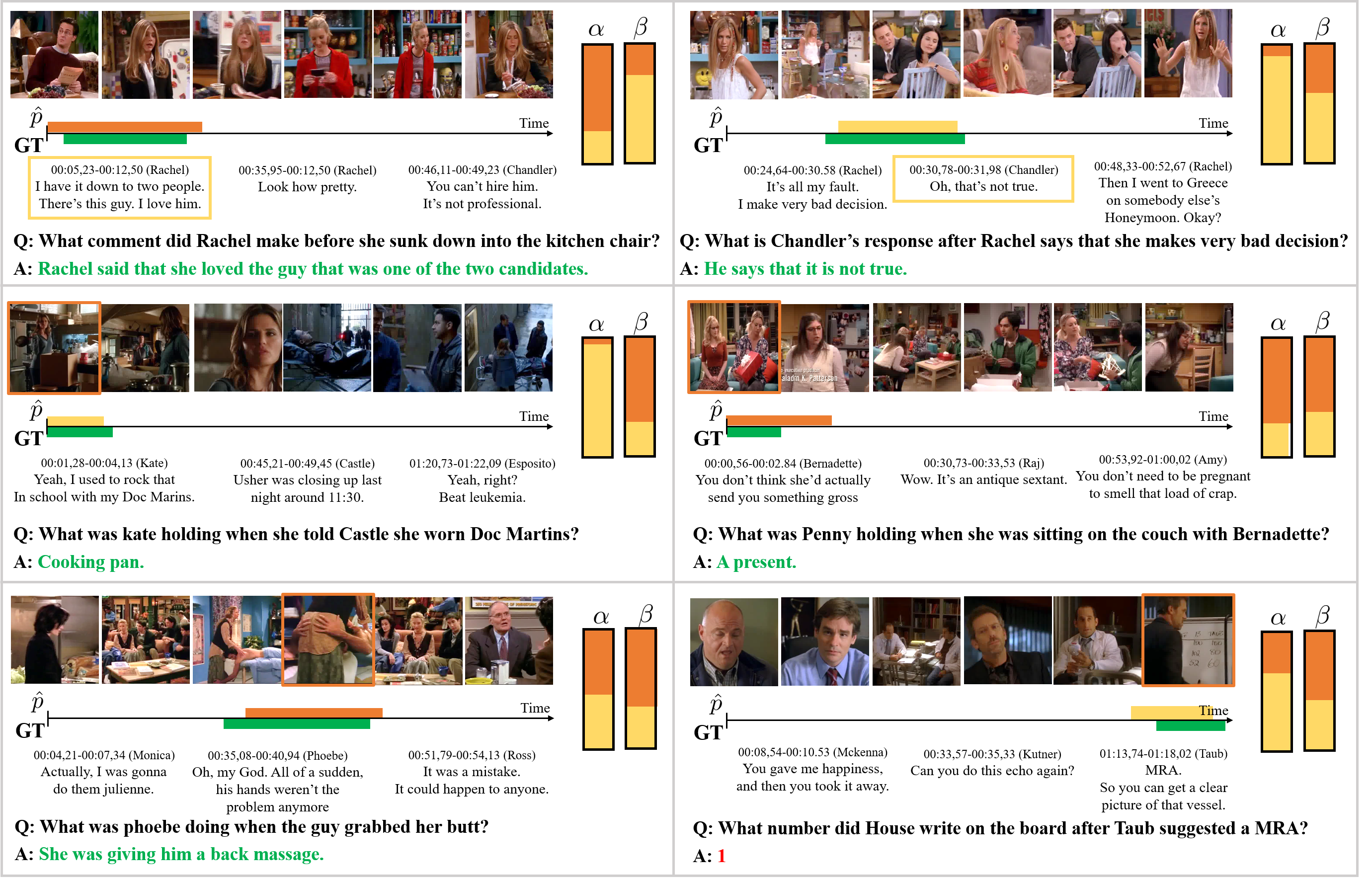}
	\caption{Visualization on the inference path of MSAN (the last example is a failure case). Each example provides MIM weights, the localized temporal moment $\hat{p}$ and ground-truth (GT) temporal moment. Video and subtitle modality are represented with orange and yellow color, respectively. The proposed MSAN dynamically modulates both modalities according to the input question.}
	\label{fig:qual1}
\end{figure*}

One observation made from Figure \ref{fig:modality} is that the accuracy on questions that require subtitle for answer prediction, i.e. $(S,V)$ and $(S,S)$ combined, is high with $86\%$ while accuracy based on video, i.e. $(V,V)$ and $(V,S)$ combined, is lower with $60\%$. This result indicates that our model does well when the answer is in the subtitle while it can do better when answer is in the video clip.

\subsubsection{Visualization of inference mechanism}
Figure \ref{fig:qual1} visualizes the inference mechanism of MSAN with selected samples from TVQA validation set. Each example is provided with MIM weights $\alpha, \beta$, localized MoI $\hat{p}$, ground-truth (GT) temporal moment and the final answer choice.
Each sample requires different combination of modalities (e.g. in the first example: video to localize and subtitle to answer, in the third example: subtitle to localize and video to answer, ...) to correctly localize and answer.
We visualize the use of video and subtitle modality using orange and yellow color, and represent it on the localized moment and key sentence or video shot.
In the first example, the model  utilizes video modality to localize the moment ($\alpha > 0.5$), and then uses subtitle modality to predict the answer ($\beta < 0.5$). As such, MSAN successfully modulates the output of the temporal moment localizer and the answer predictor with two sets of modulation weights $\alpha$ and $\beta$.
The last example shows one failure case. MSAN succeeded in localizing the key moment by using the subtitle modality ($\alpha < 0.5$). However, the model fails to predict the correct answer (i.e. $60$) as the visual concept and action concept features are insufficient in capturing textual cues in the video.

\section{Conclusion}
In this paper, we first propose to decompose MVQA into two sub-tasks: (1) localization of temporal moment relevant to the question, and (2) prediction of the correct answer based on the localized moment.
Our fundamental motivation is that the modality required for temporal localization may be different from that for the answer prediction.
To this end, the proposed Modality Shifting Attention Network (MSAN) includes two main components for each sub-task: (1) moment proposal network (MPN) that finds a specific temporal moment, and (2) heterogeneous reasoning network (HRN) that predicts the answer using multi-modal attention mechanism.
We also propose Modality Importance Modulation (MIM) to enable the modality shifting for MPN and HRN. 
MSAN showed state-of-the-art performance on TVQA dataset by achieving 71.13\% test set accuracy.

{\small
\bibliographystyle{ieee_fullname}
\bibliography{egbib}
}

\end{document}